\documentclass{article} 
\usepackage[preprint]{colm2026_conference}

\usepackage{microtype}
\usepackage{hyperref}
\usepackage{url}
\usepackage{booktabs}
\usepackage{graphicx}
\usepackage{amsmath, amssymb}
\usepackage[most]{tcolorbox}
\usepackage{enumitem}
\usepackage{subcaption}
\usepackage[export]{adjustbox}
\usepackage{tikz}
\usetikzlibrary{positioning,fit}
\usepackage{tabularx, booktabs}
\usepackage{multirow}
\usepackage{multicol}

\usepackage{lineno}

\definecolor{darkblue}{rgb}{0, 0, 0.5}
\hypersetup{colorlinks=true, citecolor=darkblue, linkcolor=darkblue, urlcolor=darkblue}

\title{\includegraphics[width=1cm]{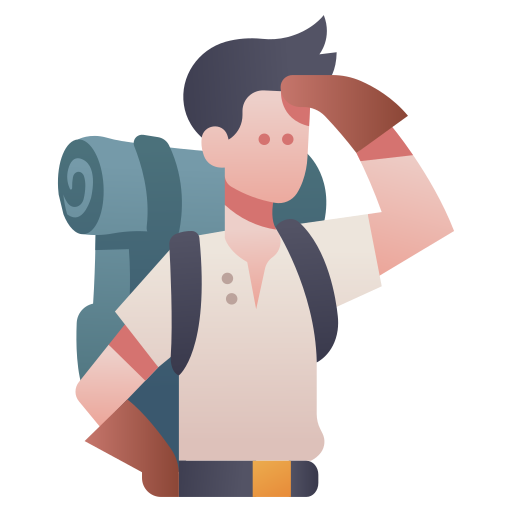}Experiences Build Characters: The Linguistic Origins and Functional Impact of LLM Personality}

\author{Xi Wang \\ 
University of Sheffield\\
Sheffield, UK \\
\texttt{xi.wang@sheffield.ac.uk} \\
\And
Mengdie Zhuang \\
University of Sheffield\\
Sheffield, UK \\
\texttt{m.zhuang@sheffield.ac.uk}
\And
Jiqun Liu \\
University of Oklahoma \\
Oklahoma, USA\\
\texttt{jiqunliu@ou.edu} \\
}

\begin{document}

\ifcolmsubmission
\linenumbers
\fi

\maketitle

\begin{abstract}
Human problem-solving is enriched by a diversity of styles and personality traits, yet the development of Large Language Models (LLMs) has largely prioritized uniform performance benchmarks that favour specific behavioural tendencies such as assertiveness. To investigate how diverse experiences shape machine personality and influence problem-solving, this study employs continued pre-training to expose models to domain-specific texts in an unsupervised manner, simulating the accumulation of experience. By adapting the Big Five framework via the Machine Personality Inventory (MPI), we quantify the personality traits of these model variants and analyse their relationship to linguistic style and reasoning behaviour. The findings reveal that model competence is \textit{bimodal}, peaking at "Expressive Generalists" and "Suppressed Specialists," while identifying a "Suppression Advantage" where reduced social traits enhance complex reasoning performance. This study further establishes a causal link between training data linguistics, such as imperative frequency, and lexical diversity, providing a roadmap for "Personality Engineering".
\end{abstract}

\section{Introduction}

Humans rarely approach a problem in the same way \citep{treffinger2008understanding}. When asked to solve a puzzle, write an argument,  draft an essay, or provide suggestions, individuals may come up with responses that are equally valid but different in style. One may give a concise and definitive answer, another may carefully weigh alternatives before deciding, while yet another may offer an unconventional perspective \citep{DZURILLA2011142}. These variations \textit{reflect the diversity of human experiences and the enduring influence of personality traits}. This diversity is vital for problem-solving, allowing groups to draw on complementary strengths, encouraging creativity and curiosity, and preventing the limitation of thought that can arise from converging on a single, uniform style. In contrast, the development of large language models (LLMs) has largely prioritized certain measures of performance 
\citep{chang2024survey}, defined by benchmarks that often reward accuracy on knowledge- and reasoning-oriented tasks, such as MMLU \citep{wang2024mmlu} and HellaSwag \citep{zellers2019hellaswag}. Success on these benchmarks often privileges specific behavioural tendencies, such as assertiveness or fluency, which resemble certain human personality traits. For instance, a model that confidently provides direct answers may score higher than a model that responds cautiously or considers multiple alternatives, even if both are equally plausible in human communication. Instruction tuning further amplifies this trend by imposing a uniform, "helpful" persona, suppressing variation in reasoning and communication styles. As a result, contemporary LLMs risk converging towards a narrow behavioural profile, optimised for leaderboards rather than producing human-like intellectual diversity.

Nevertheless, systematic investigations into how diverse experiences shape personality traits in LLMs and how these traits influence problem-solving styles and performances remain limited~\citep{he2025investigating, chen2025mitigating}. To address this gap, we employ continued pre-training to encode diversified “experiences” into LLMs and model their resulting personality traits. Unlike instruction tuning or reinforcement learning with human feedback \citep{ouyang2022training}, which guides models toward fixed behavioral traits through labeled or preference data, continued pretraining \citep{ke2023continual} exposes models to domain-specific 
texts in an unsupervised manner. This allows them to internalise linguistic style, topical focus, and implicit reasoning patterns without externally imposed objectives. Analogous to how human personalities form through accumulated experiences, 
continued pretraining on diverse domains induces differentiated linguistic strategies and cognitive dispositions in LLMs.
More than acting as an unsupervised adaptation method, it also provides a framework to explore how \textit{experience-driven} diversity gives rise to varied model behaviours and performances.

After continued pretraining, we obtain variant LLMs with distinct exposure histories. To quantify their personality traits, we adopt the Machine Personality Inventory (MPI) \citep{jiang2023evaluating}, which adapts the Big Five psychometric framework \citep{de2000big} into a multiple-choice test for LLMs. By prompting each model with MPI tests, we derive scores on five systematic personality traits (i.e., Extraversion, Agreeableness, Conscientiousness, Neuroticism, Openness) and analyse how personality profiles vary depending on the models’ exposure histories. 
This enables us to link experience-induced variation to measurable differences in \textit{linguistic style}, \textit{reasoning behaviour}, and \textit{problem-solving tendencies}. We further conduct analysis into linguistic signals (e.g., sentence complexity and sentiment polarity) to capture potential links with model variants' performance. 

The experimental results show success peaks at Expressive Generalists and Suppressed Specialists, while intermediate profiles suffer from dissonance. Notably, we identify a "Suppression Advantage" where reduced social traits enhance performance on complex reasoning tasks. Furthermore, the syntactic analysis of the corpus reveals that these traits are direct artefacts of training data linguistics, such as imperative frequency driving extraversion. In sum, the contributions of this study include: (1) \textbf{Framework}: We validate that domain-specific pretraining induces distinct, measurable personality traits in LLMs, proving that "experiences build character." (2) \textbf{Empirical Insight}: We demonstrate the bimodal nature of model competence and the functional advantage of personality suppression for rigorous analytical tasks. (3) \textbf{Mechanism}: We establish the causal link between corpus linguistics (e.g., Imperative Ratio, Lexical Diversity) and the emergence of specific personality traits, providing a roadmap for "Personality Engineering."

\section{Related Work}

\paragraph{Problem-Solving and Machine Psychology}
Human problem-solving has been extensively studied in psychology and cognitive science. Simon’s theory of bounded rationality showed that individuals satisfice rather than optimize, relying on heuristics under constraints of information and cognition~\citep{simon1957behavioral}. Kahneman and Tversky later demonstrated that judgments follow systematic heuristics and biases, highlighting the dual processes of intuitive and deliberative reasoning~\citep{tversky1974judgment, kahneman2011thinking}. Beyond these universal mechanisms, personality has been shown to play a central role in shaping problem-solving effectiveness. Traits such as openness and conscientiousness predict creativity, persistence, and strategic flexibility, while neuroticism often correlates with less adaptive outcomes~\citep{de2000big, treffinger2008understanding, DZURILLA2011142}. Early trait-based studies also linked personality dimensions to achievement and innovation~\citep{cattell1968prediction}. Collectively, this body of work highlights that human problem-solving is not only bounded by cognitive limitations but also enriched by individual differences, which contribute to the diversity of reasoning strategies and collective adaptability.

LLMs have introduced parallel questions about machine problem-solving. Standard evaluations emphasize accuracy and benchmark performance, rewarding confident and fluent outputs while discouraging reasoning diversity~\citep{zellers2019hellaswag, wang2024mmlu}. Recent large-scale studies suggest that reasoning abilities can emerge abruptly at scale, with chain-of-thought reasoning and analogical inference appearing only when models surpass certain parameter thresholds~\citep{wei2022emergent}. Reports on GPT-4 further illustrate the breadth of these emergent capabilities, and document problem-solving across mathematics, law, and logic that rivals human benchmarks~\citep{bubeck2023sparks}. Meanwhile, research shows that alignment methods, while improving safety and controllability, may reduce behavioural richness, a phenomenon described as an alignment tax~\citep{ouyang2022training}. Cognitive and behavioural analyses further indicate that LLMs reproduce heuristics, biases, and elements of theory of mind, though inconsistently~\citep{binz2023using, he2025investigating, kosinski2024evaluating}. These insights point to the emergence of a \textit{Machine Psychology}~\citep{hagendorff2023machine, chen2025mitigating} perspective: LLMs, like humans, exhibit bounded and bias-prone reasoning, shaped by scale, training, and alignment. However, unlike in human research where dispositional diversity is recognized as a strength, little is known about how variation in LLMs translates into distinct problem-solving styles or how diversity in personality characteristics might be harnessed constructively. This gap motivates our study, which systematically examines the relationship between induced personality traits and LLMs' behaviour, bridging psychology and computational modelling.

\paragraph{Personality of Human and Machine}
Personality refers to enduring patterns of cognition, affect, and behavior that distinguish individuals and shape how they interact with the world~\citep{matthews2003personality}. The Five-Factor Model has provided a robust and replicable framework for describing these patterns across contexts~\citep{john1999big, costa1999five}. Traits such as openness and conscientiousness have been linked to creativity, persistence, and goal-directed behavior, whereas agreeableness and extraversion influence social adaptability and cooperation~\citep{kain2000social}. Personality differences affect cognitive flexibility and problem-solving style by modulating tolerance for ambiguity, attentional control, and motivational orientation. In group or organizational settings, diversity in personality composition has been associated with improved collaboration and innovation through the combination of complementary cognitive strategies~\citep{judge2013hierarchical}. These findings demonstrate that stable dispositional variation enhances adaptive reasoning, and thus offer a psychological foundation for studying individual differences in human minds and machines.

Recent research on LLMs has extended these ideas into computational settings, examining whether and how such models exhibit personality-like regularities. \citet{jiang2023evaluating} demonstrated that prompting or fine-tuning can induce measurable and interpretable trait profiles in LLMs that align with the Big Five dimensions, and that human evaluators consistently recognize these differences in generated text. \citet{tseng2024two} surveyed work on persona construction and found two dominant paradigms: role-playing, in which models are conditioned to act as specific personas, and personalization, in which models adapt to user characteristics over time. \citet{hu2024quantifying} quantified the “persona effect,” showing that personality conditioning slightly alters reasoning and affective expression in dialogue generation but that consistency across turns remains limited. \citet{bhandari2025can} further examined conversational stability and found that LLMs struggle to maintain coherent persona expression throughout extended interactions. \citet{caron2022identifying} found that personality traits of language models shift in response to and mimicking other personalities present in the context. Findings from \citet{han2024psydial} suggest that fine-tuning for personality expression can influence linguistic style and self-referential reasoning, suggesting that personality cues may serve as a controllable factor in LLM behaviour. These studies jointly show that personality modelling in LLMs is both measurable and malleable, creating opportunities to investigate how induced trait profiles shape reasoning diversity, hidden biases and manipulation risks, and alignment with human~\citep{liu2026bounded}. However, the stability, transferability, and functional impact of such induced personalities remains underexplored. Investigating these dimensions can help integrate established psychological theory with computational modelling to enhance adaptability in generative AI systems.

\section{Methodology}

In this study, we investigate how current large language models (LLMs) with distinct exposure histories approach problem-solving tasks and how their latent personality traits influence their performance. Formally, let $\mathcal{M} = \{m_1, m_2, ..., m_N\}$ denote a set of LLM variants obtained through continued pretraining on diverse corpora ($D$). Each model $m_i$ is evaluated on a set of problem-solving tasks $\mathcal{T} = \{t_1, t_2, ..., t_K\}$, where each task $t_k$ is designed to assess reasoning, analytical, or decision-making ability. The performance of a model $m_i$ on a given task $t_k$ is represented as $P_{i,k}$, and the overall performance of $m_i$ across all tasks is denoted by $P_i = \{P_{i,1}, P_{i,2}, ..., P_{i,K}\}$.

To examine personality traits, we associate each model $m_i$ with a personality vector $\mathbf{T}_i = \{T_{i,1}, T_{i,2}, ..., T_{i,5}\}$, where each element corresponds to one of the five major personality dimensions defined in the Machine Personality Inventory (MPI)~\citep{jiang2023evaluating}, adapted from the Big Five framework~\citep{de2000big}. The trait vector $\mathbf{T}_i$ captures the model’s behavioural and linguistic characteristics along \textit{Extraversion}, \textit{Agreeableness}, \textit{Conscientiousness}, \textit{Neuroticism}, and \textit{Openness}.

This study aims to investigate the relationship between the model’s personality vector $\mathbf{T}_i$ and its task performance $P_i$. Specifically, we aim to estimate whether certain traits correlate positively or negatively with different categories of problem-solving ability. By systematically analysing the mapping \(f\!:\!\mathbf{T}_i \!\rightarrow\! P_i\), we seek to understand how experience-driven personality differences in LLMs influence their reasoning, communication, and decision-making styles. Hence, we develop a two-stage methodology: first, constructing LLM variants through continued pretraining, then quantifying their personality traits using the Machine Personality Inventory (MPI) \citep{jiang2023evaluating}, and analysing the relationships between personality traits and problem-solving performance.

\subsection{Continued Pretraining}

To simulate diverse experiential exposure in large language models (LLMs), we adopt the \textbf{continued pretraining} strategy, or \textit{domain-adaptive pretraining (DAPT)}~\citep{gururangan2020don}. This approach extends unsupervised language modelling to new corpora, enabling each model to internalise domain-specific linguistic and conceptual patterns without supervision. In this context, continued pretraining serves as a computational analogue of \textit{experience exploration}, where different textual domains act as distinct learning experiences that shape the model’s behaviour and personality traits.

Formally, given a pretrained base model $m_0$ and a new corpus $D_j$ representing a specific experiential domain, the objective is to maximise the standard causal language modelling likelihood:

\small
\[
\mathcal{L}_{\mathrm{CausalLM}} = - \sum_{t=1}^{T} \log P_\theta (x_t \mid x_{<t}),
\]
\normalsize
where $x_t$ denotes the $t$-th token in the input sequence and $\theta$ are the model parameters initialised from $m_0$. Through this process, we obtain a domain-adapted model $m_j$ that preserves the general linguistic competence of $m_0$ while internalising additional stylistic and conceptual patterns specific to $D_j$.
This unsupervised adaptation differs fundamentally from supervised fine-tuning or reinforcement learning with human feedback (RLHF), which explicitly optimise for task-specific or preference-aligned behaviour. In contrast, continued pretraining relies purely on exposure to text distributions, mirroring how human personalities and cognitive styles develop through accumulated experience rather than explicit instruction. In this study, each corpus $D_j$ corresponds to a distinct "experiential domain" (e.g., \textit{biomedical}, \textit{literary}, \textit{legal}), yielding a set of LLM variants $\mathcal{M} = \{m_1, m_2, ..., m_n\}$ that embody diverse learned experiences. These models serve as the foundation for analysing how variation in textual exposure shapes the emergence of personality traits and their influence on problem-solving behaviour.

\subsection{Personality Trait Evaluation and Analysis}\label{ssec:personality_eval}
Each model $m_i \!\in\! \mathcal{M}$ is assessed using the Machine Personality Inventory (MPI)~\citep{jiang2023evaluating}, which adapts the Big Five framework~\citep{de2000big} into a multiple-choice test format suitable for LLMs. By following \citet{jiang2023evaluating} and prompting each model with MPI items, we derive a five-dimensional trait vector $\mathbf{T}_i = \{T_{i,1}, T_{i,2}, T_{i,3}, T_{i,4}, T_{i,5}\}$ corresponds to five well-established psychological dimensions:

\small
\begin{tcolorbox}[colback=gray!5,colframe=gray!40,boxrule=0.5pt,arc=2pt,left=6pt,right=6pt,top=4pt,bottom=4pt]
\textbf{Openness:} artistic, curious, imaginative, and intellectually explorative.\\[3pt]
\textbf{Conscientiousness:} organised, responsible, reliable, and goal-oriented.\\[3pt]
\textbf{Extraversion:} energetic, assertive, enthusiastic, and socially expressive.\\[3pt]
\textbf{Agreeableness:} kind, cooperative, empathetic, and considerate.\\[3pt]
\textbf{Neuroticism:} anxious, sensitive, emotionally unstable, and easily stressed.
\end{tcolorbox}
\normalsize
\noindent The resulting trait scores capture the stable personality tendencies expressed by each model variant.

Next, to study how personality traits influence model behaviour, we quantify associations between personality and performance using two independent procedures.

\textbf{Single-trait.}
For each trait $j \in \{1,\dots,5\}$, we compute the Pearson correlation between the trait score and performance:

\small
\[
r_j \;=\; \operatorname{corr}(T_{i,j},\, P_i).
\]
\normalsize

\textbf{Paired-trait (weighted linear combination).}
For each ordered pair $(j,k)$ of personality trait scores, with $j \neq k$, we fit a bivariate linear model with the two trait scores as predictors:

\small
\[
P_i \;=\; \beta_0 \;+\; \beta_j\,T_{i,j} \;+\; \beta_k\,T_{i,k} \;+\; \varepsilon_i,
\]
\normalsize
where $(\beta_j,\beta_k)$ are estimated by ordinary least squares on the matrix
$X=[\,T_{i,j}\;\; T_{i,k}\,]$. 
The fitted coefficients quantify the partial effect of each trait while controlling for the other. 
Optionally, we form the weighted linear combination
$S_{i,jk}=\beta_j T_{i,j}+\beta_k T_{i,k}$ and report 
$\rho_{jk}=\operatorname{corr}(S_{i,jk},\,P_i)$ as a summary of the joint contribution without multiplicative interaction terms. This formulation preserves the interpretability of each personality dimension by weighting their influence according to the fitted coefficients, rather than arbitrarily combining them through multiplication or direct summation. Such multiplicative or additive operations are psychologically invalid, as the Big Five traits are empirically orthogonal constructs that represent independent behavioural tendencies rather than compositional factors \citep{saucier2002orthogonal}. The weighted linear combination, therefore, provides a more principled way to capture the coordinated influence of two traits while respecting the theoretical independence among dimensions.

\section{Experimental Setup}

We apply continued pretraining to the publicly available Llama-3-8B, using a copyright-filtered version of The Pile dataset \citep{gao2020pile}. Specifically, we remove all known copyrighted subsets (e.g., Books3, OpenSubtitles) to respect author rights while preserving domain diversity. Table \ref{tab:datasets_roles_vertical} in the Appendix presents the list of subsets (e.g., ArXiv and FreeLaw) with their corresponding descriptions. These datasets span diverse domains, including academic, legal, biomedical, literary, technical, and general web, which provide the experiential variety needed to induce differentiated personality and behavioural patterns in LLM variants. To ensure fair comparison under limited compute, we standardised the token budget across domains based on the smallest corpus, \textit{Gutenberg}. Each corpus was tokenised into 512-length sequences, from which we randomly sampled the number required to match 68,551,839 tokens (the total from \textit{Gutenberg}). This “calculated token” sampling guarantees equal data volume across variants to avoid unexpected compounding factors.

Next, to develop meaningful character with experience exposure and construct coherent domain combinations, we used GPT-5 \citep{openai2025gpt5} as a creative assistant to suggest dataset groupings aligned with persona roles (e.g., “PubMed Central + NIH Exporter” for a biomedical expert). The proposed combinations were manually reviewed and adopted to ensure semantic compatibility and thematic coherence across the training corpora. This results in 11 characters, and their corresponding use of pretraining data subsets is presented in Table \ref{tab:datasets_roles_vertical}. For the continued pretraining, it is done with a learning rate of $1e^{-5}$, a warm-up ratio of 0.1 using a cosine scheduler, weight decay of 0.01, and mixed precision. Each domain model is trained for 1 epoch with a batch size of 10 per device. For performance evaluation of our LLM variants, we follow many common practices and include various standard language modelling benchmarks,  \textbf{Massive Multitask Language Understanding (MMLU)} benchmark~\citep{hendrycks2021measuring}, and its enhanced successor \textbf{MMLU-Pro}~\citep{wang2024mmlu}.
MMLU assesses model performance across 57 subject areas (STEM, humanities, social sciences, law, medicine) using multiple-choice questions to test knowledge and reasoning capabilities.  
MMLU-Pro increases difficulty by expanding the number of answer options, removing trivial items, and focusing more on reasoning-centric domains.  

In this study, we aim to answer the following research questions:
\textbf{RQ1:} Do LLMs pretrained on diverse domain corpora exhibit performance advantages in task domains aligned with their continued pretraining resources?  
\textbf{RQ2:} Do personality trait scores evaluated via MPI correlate with performance across benchmarks, including rich tasks about knowledge and reasoning?  

\section{Result Analysis}

\begin{table*}[t]
\centering
\small
\resizebox{\textwidth}{!}{
\begin{tabular}{lrrrrrrr|rrrrrrr}
\toprule
{} & \multicolumn{7}{c}{MMLU} & \multicolumn{7}{c}{MMLU-Pro} \\
\cmidrule(lr){2-8} \cmidrule(lr){9-15}
Model & STEM & Health & Soc. Sci. & Law & Humanit. & Psych. & Other &
STEM & Health & Soc. Sci. & Law & Humanit. & Psych. & Other \\
\midrule
Llama-3-8b               & 0.5302 & 0.6975 & 0.7124 & 0.7389 & 0.6821 & 0.7745 & 0.8186 & 0.3039 & 0.4920 & 0.3949 & 0.2189 & 0.3959 & 0.5401 & 0.4123 \\
\hline
Literary Classicist      & 0.4843 & 0.6444 & 0.6656 & 0.7172 & 0.6534 & 0.7440 & 0.7740 & 0.2716 & 0.4761 & \textbf{0.3825} & \textbf{0.2443} & 0.3463 & \textbf{0.5263} & 0.3669 \\
Inventive Technologist   & 0.5067 & 0.6835 & 0.6843 & 0.7219 & 0.6720 & 0.7451 & 0.8084 & 0.2469 & 0.4635 & 0.3183 & 0.2089 & 0.3565 & 0.4937 & 0.3788 \\
Technical Communicator   & \textbf{0.5165} & 0.6849 & 0.6851 & \textbf{0.7395} & \textbf{0.6854} & 0.7629 & 0.8072 & 0.2743 & \textbf{0.4802} & 0.3533 & 0.2134 & 0.3652 & 0.5150 & \textbf{0.4015} \\
Business Advisor         & 0.5073 & \textbf{0.6869} & 0.6821 & 0.7274 & 0.6764 & 0.7455 & 0.8033 & 0.2395 & 0.4637 & 0.3438 & 0.1944 & \textbf{0.3879} & 0.4975 & 0.3820 \\
Health Advisor           & 0.4555 & 0.6214 & 0.6280 & 0.6837 & 0.6265 & 0.6790 & 0.7484 & 0.2309 & 0.4158 & 0.3043 & 0.1689 & 0.3149 & 0.4461 & 0.3268 \\
Scientific Scholar       & 0.4648 & 0.6086 & 0.6447 & 0.6584 & 0.6368 & 0.6724 & 0.7510 & 0.2449 & 0.4148 & 0.2990 & 0.1762 & 0.3226 & 0.4524 & 0.3344 \\
Scientific Mathematician & 0.4963 & 0.6763 & 0.6816 & 0.7052 & 0.6755 & 0.7394 & \textbf{0.8174} & 0.2372 & 0.4523 & 0.3217 & 0.1944 & 0.3637 & 0.4987 & 0.3658 \\
Legal Analyst            & 0.4939 & 0.6740 & 0.6800 & 0.7019 & 0.6595 & 0.7387 & 0.8008 & 0.2466 & 0.4581 & 0.3428 & 0.2035 & 0.3787 & 0.5063 & 0.3929 \\
Biomedical Expert        & 0.5031 & 0.6756 & \textbf{0.7035} & 0.7339 & 0.6835 & 0.7579 & 0.8046 & \textbf{0.2756} & 0.4630 & 0.3319 & 0.1880 & 0.3735 & 0.4925 & 0.3994 \\
Patent Strategist        & 0.5120 & 0.6849 & 0.6964 & 0.7272 & 0.6830 & \textbf{0.7704} & 0.7867 & 0.2658 & 0.4573 & 0.3429 & 0.1889 & 0.3736 & 0.4975 & 0.3896 \\
Cultural Scholar         & 0.4627 & 0.6210 & 0.6367 & 0.6709 & 0.6283 & 0.6990 & 0.7612 & 0.2599 & 0.3989 & 0.3219 & 0.2089 & 0.3504 & 0.4561 & 0.3333 \\
\bottomrule
\end{tabular}
}
\caption{
Averaged accuracy of LLM variants across unified domain groups on \textbf{MMLU} and \textbf{MMLU-Pro}. Values are grouped by domain mappings (Appendix Table~\ref{tab:mmlu_mapping}); bold numbers indicate the best-performing variant per domain.}
\label{tab:mmlu_mmlupro}
\end{table*}

\paragraph{Experience Exposure on Domain-Specific Problem Solving} To address RQ1, we examine whether large language models (LLMs) with distinct experiential exposure, achieved through continued pretraining on domain-specific corpora, demonstrate improved performance in corresponding task domains across MMLU and MMLU-Pro. Both benchmarks cover diverse categories spanning STEM, health, social science, law, humanities, and psychology. However, MMLU contains 57 fine-grained task categories, while MMLU-Pro includes only 14. To ensure consistent comparison between the two datasets, we grouped their task categories into unified domain clusters, as detailed in Appendix Table~\ref{tab:mmlu_mapping}.

Table~\ref{tab:mmlu_mmlupro} presents the averaged accuracies across these unified domains. The comparative patterns across MMLU and MMLU-Pro indicate that certain variants, such as the Technical Communicator, achieve consistently strong performance across multiple domains, even beyond those directly related to their pretraining corpora. Conversely, \textit{domain alignment alone does not directly guarantee domain-specific gains}. For instance, the Patent Strategist pretrained on FreeLaw shows improvements in legal reasoning, while the Scientific Scholar and Scientific Mathematician underperform in STEM categories relative to other variants. This inconsistency implies that continued pretraining may enhance stylistic or lexical adaptation more than it develops deeper reasoning competence. The learning signals embedded in domain corpora may shape the model’s communication style and response tendencies, effectively altering its “personality”, without guaranteeing improvements in analytical or inferential ability. Interestingly, variants such as the Technical Communicator exhibit superior cross-domain robustness, performing well even on tasks semantically unrelated to their training resources. This raises the possibility that certain personality-like tendencies, emerging from experience exposure, contribute to problem-solving adaptability rather than domain-specific expertise.

In sum, continued pretraining induces measurable behavioural differences among LLM variants, yet domain relevance alone does not fully explain performance variations. Some models demonstrate broader generalisation despite limited task relevance, suggesting that latent personality traits, shaped through exposure to specific communicative environments, may influence how models approach reasoning and decision-making. This observation motivates our further investigation in \textbf{RQ2}, where we systematically examine the relationship between personality traits and problem-solving performance across tasks.

\begin{table*}[t]
\centering
\small
\resizebox{0.75\textwidth}{!}{%
\begin{tabular}{l*{5}{cc}}
\toprule
\multirow{2}{*}{Model}
  & \multicolumn{2}{c}{Openness}
  & \multicolumn{2}{c}{Conscientiousness}
  & \multicolumn{2}{c}{Extraversion}
  & \multicolumn{2}{c}{Agreeableness}
  & \multicolumn{2}{c}{Neuroticism} \\
\cmidrule(lr){2-3}\cmidrule(lr){4-5}\cmidrule(lr){6-7}\cmidrule(lr){8-9}\cmidrule(lr){10-11}
 & Score & $\sigma$ & Score & $\sigma$ & Score & $\sigma$ & Score & $\sigma$ & Score & $\sigma$ \\
\midrule
Meta-Llama-3-8B & 3.13 & 1.44 & 3.25 & 1.34 & 3.33 & 1.32 & 3.07 & 1.39 & 3.14 & 1.41 \\
\midrule
literary\_classicist & 3.12 & 1.14 & 3.19 & 1.33 & 3.16 & 1.19 & 3.03 & 1.24 & 2.75 & 1.15 \\
inventive\_technologist & 2.93 & 1.12 & 3.13 & 1.14 & 2.91 & 1.17 & 3.11 & 1.14 & 2.87 & 1.14 \\
patent\_strategist & 3.13 & 1.41 & 3.30 & 1.37 & 3.02 & 1.42 & 3.01 & 1.40 & 2.84 & 1.45 \\
cultural\_scholar & 2.88 & 1.33 & 2.93 & 1.30 & 2.93 & 1.32 & 3.01 & 1.30 & 3.06 & 1.37 \\
technical\_communicator & 3.00 & 1.22 & 2.99 & 1.18 & 2.91 & 1.18 & 2.88 & 1.22 & 2.82 & 1.12 \\
business\_advisor & 2.84 & 1.18 & 3.22 & 1.28 & 3.10 & 1.30 & 3.08 & 1.28 & 2.99 & 1.20 \\
health\_advisor & 3.27 & 1.40 & 3.46 & 1.43 & 3.16 & 1.36 & 3.02 & 1.41 & 3.15 & 1.52 \\
scientific\_scholar & 3.14 & 1.47 & 3.13 & 1.48 & 3.06 & 1.35 & 2.98 & 1.40 & 2.82 & 1.39 \\
scientific\_mathematician & 2.95 & 1.34 & 2.99 & 1.29 & 2.99 & 1.32 & 3.14 & 1.43 & 2.98 & 1.43 \\
legal\_analyst & 2.77 & 1.09 & 3.10 & 1.07 & 2.96 & 1.13 & 2.86 & 1.11 & 2.85 & 1.10 \\
biomedical\_expert & 3.05 & 1.58 & 3.06 & 1.49 & 3.13 & 1.47 & 3.05 & 1.52 & 3.03 & 1.46 \\
\bottomrule
\end{tabular}}
\caption{
LLMs’ personality analysis based on \textbf{1k-item MPI}. 
Scores are means, and $\sigma$ are standard deviations across answered items. 
The five personality dimensions are: 
\textbf{Openness} (artistic, curious, imaginative, original), 
\textbf{Conscientiousness} (efficient, organised, reliable, responsible), 
\textbf{Extraversion} (active, assertive, outgoing, talkative), 
\textbf{Agreeableness} (appreciative, kind, sympathetic), and 
\textbf{Neuroticism} (anxious, tense, worrying).
}
\label{tab:mpi_1k}
\end{table*}

\begin{figure*}[t]
    \centering
    \begin{subfigure}[b]{0.15\textwidth}
        \centering
        \includegraphics[width=\textwidth]{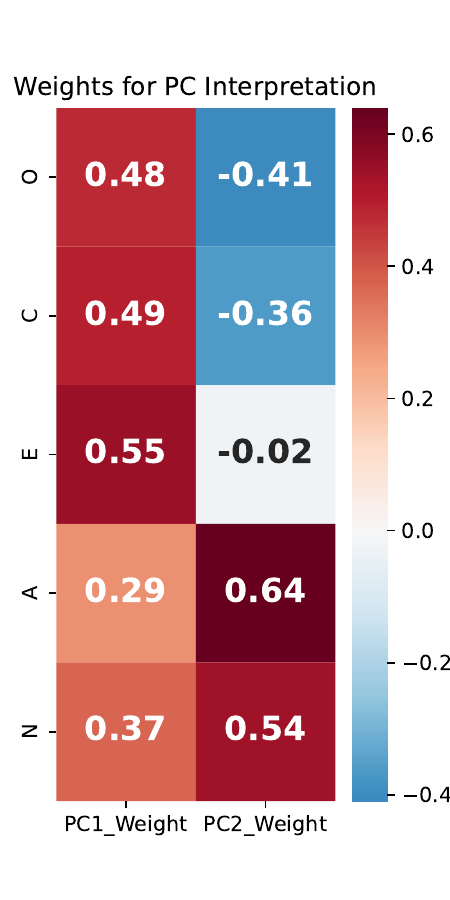}
        \caption{\textbf{Latent Axis} defines PC1 and PC2.}
        \label{fig:pca_loadings}
    \end{subfigure}
    \hfill 
    \begin{subfigure}[b]{0.4\textwidth}
        \centering
        \includegraphics[width=\textwidth]{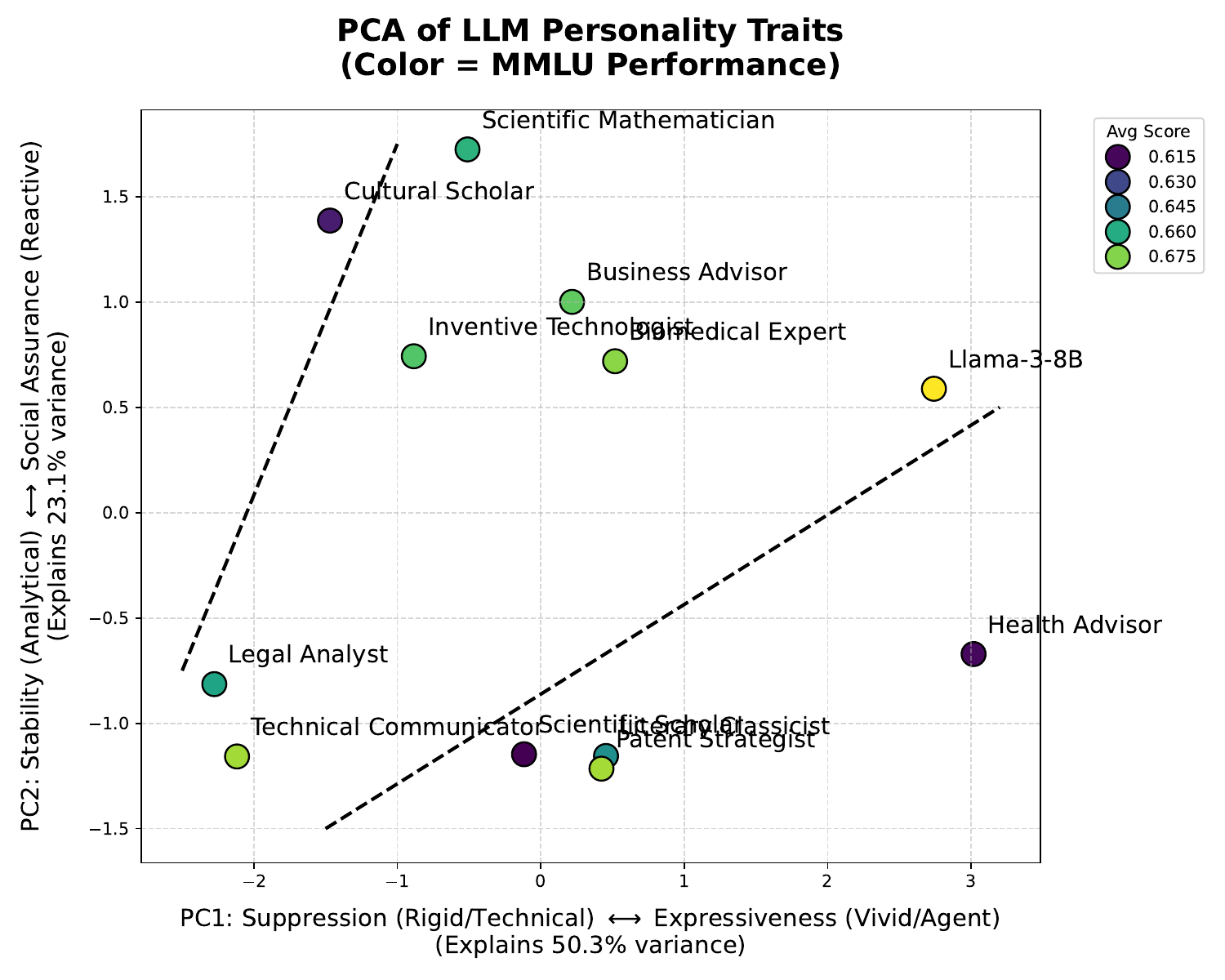}
        \caption{\textbf{MMLU Performance.}}
        \label{fig:pca_mmlu}
    \end{subfigure}
    \hfill 
    \begin{subfigure}[b]{0.4\textwidth}
        \centering
        \includegraphics[width=\textwidth]{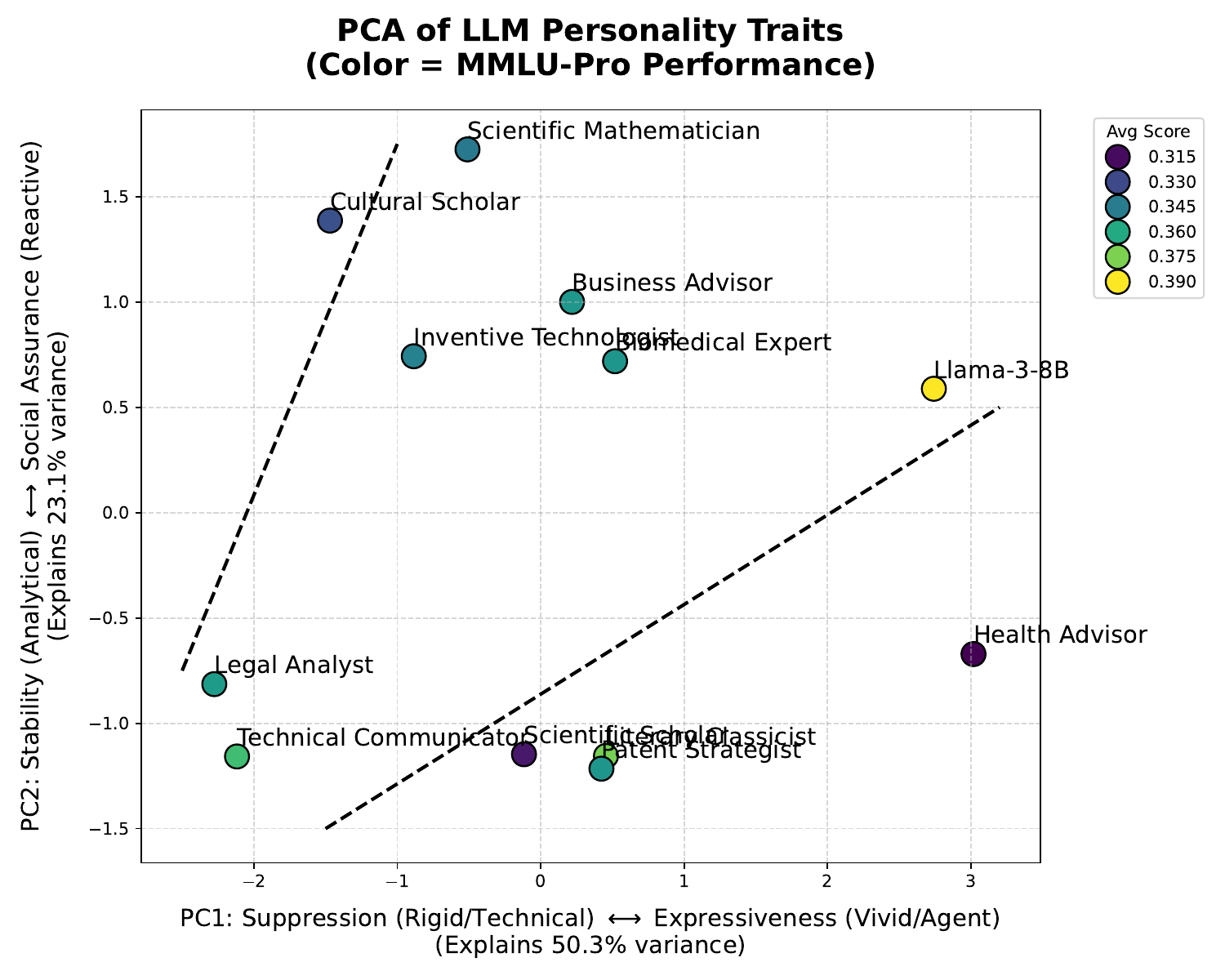}
        \caption{\textbf{MMLU-Pro Performance.}  }
        \label{fig:pca_mmlu_pro}
    \end{subfigure}
    
    \caption{\textbf{The Polarization of Competence.} 
    (a) Latent personality space is defined by Expressiveness and Social Assurance. 
    (b-c) Performance is maximized when dimensions are aligned. Black boundary lines in (b) and (c) illustrate a "Congruence Zone": models succeed either as Suppressed/Stable Tools or Expressive/Social Agents (in between two lines). }
    \label{fig:main_pca_analysis}
\end{figure*}

\begin{figure*}[t]
    \centering
    \begin{subfigure}[b]{0.48\textwidth}
        \centering
        \includegraphics[width=\textwidth]{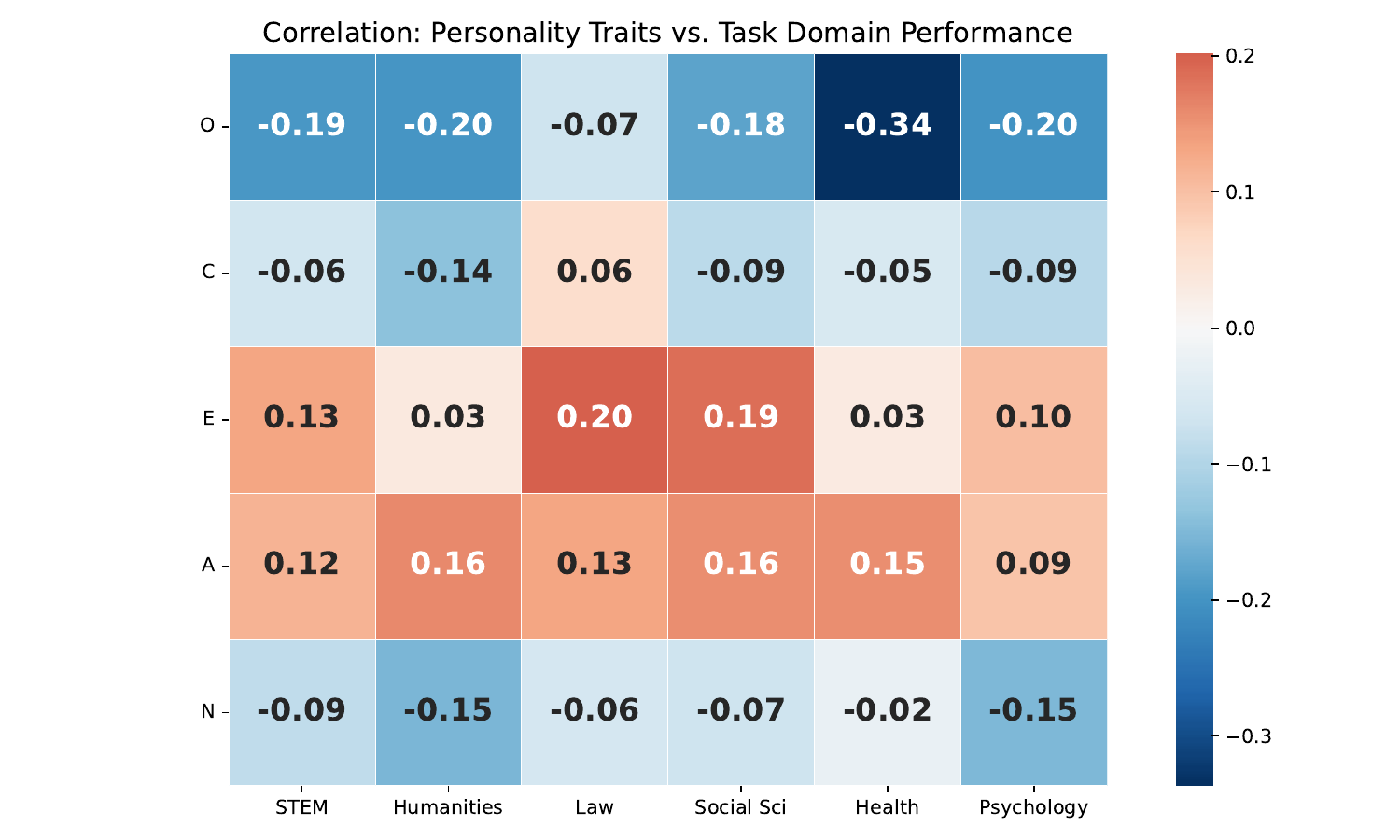}
        \caption{\textbf{MMLU Correlations (Standard Difficulty).}}
        \label{fig:corr_mmlu}
    \end{subfigure}
    \hfill
    \begin{subfigure}[b]{0.48\textwidth}
        \centering
        \includegraphics[width=\textwidth]{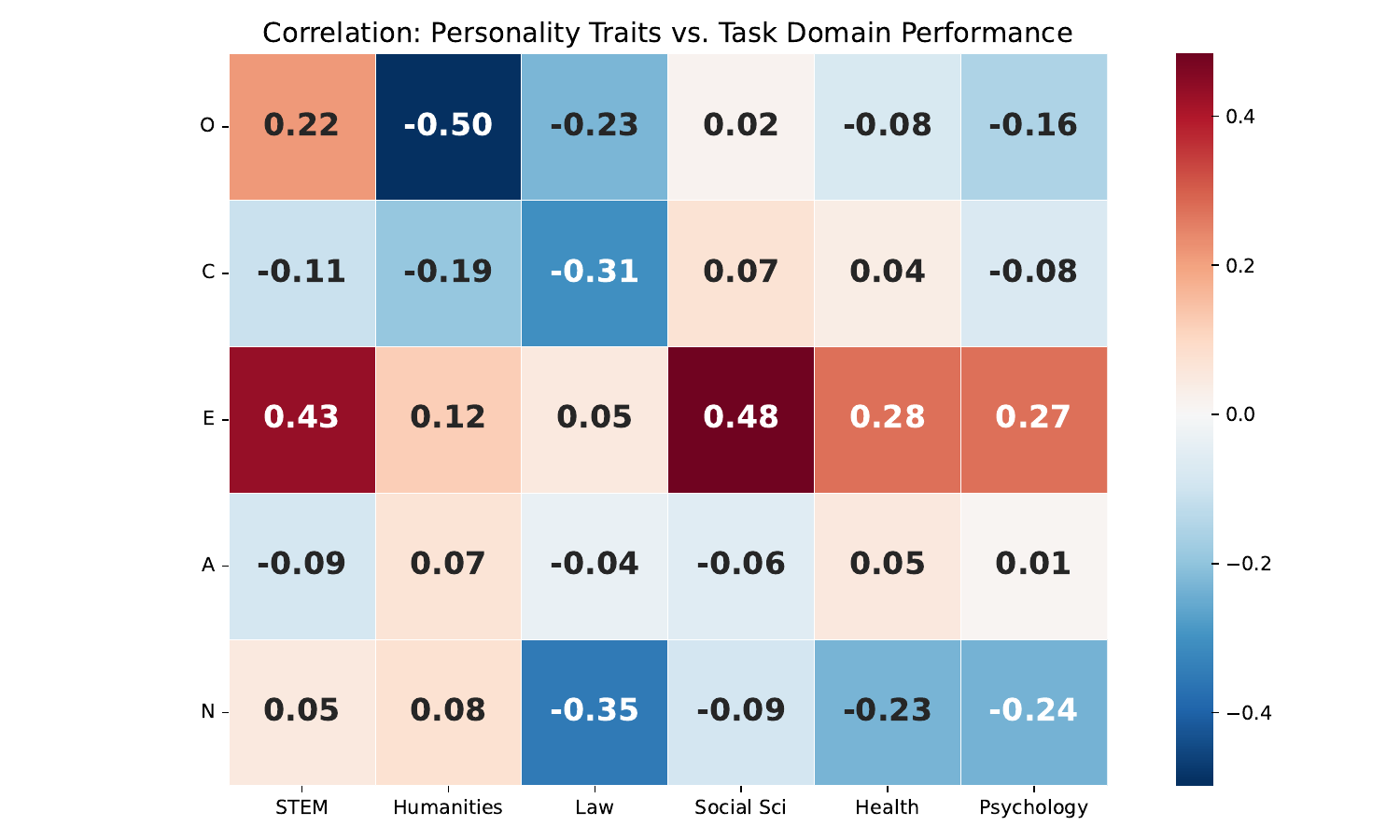}
        \caption{\textbf{MMLU-Pro Correlations (High Difficulty).}}
        \label{fig:corr_mmlu_pro}
    \end{subfigure}
    
    \caption{The Correlation of Personality Effects under Complexity.}
    \label{fig:correlation_comparison}
\end{figure*}

\paragraph{Personality Traits and Problem-Solving Performance}
To address RQ2, we examine the structural organisation of personality traits. By projecting the 5-dimensional MPI vectors onto a 2D latent space via Principal Component Analysis (Figure \ref{fig:main_pca_analysis}), we identify two primary axes driving model variation: Trait Expressiveness (PC1, 50.3\% variance) and Social Assurance (PC2, 23.1\% variance). Contrary to the expectation that "more personality is better", our analysis reveals a bimodal performance landscape. As shown in Figure \ref{fig:main_pca_analysis}, high-performing models cluster at the distinct extremes of the Expressiveness axis, while models in the "middle ground" (e.g., Scientific Scholar) consistently underperform.  Effective models fall into two congruent zones: the Consistent Agent (Top-Right, High Expressiveness + High Social Assurance) or the Consistent Tool (Bottom-Left, Low Expressiveness + High Stability). Deviating from this diagonal leads to failure. The Health Advisor (Bottom-Right) exhibits Personality Dissonance: it pairs High Expressiveness with Low Social Assurance/High Rigidity. Psychologically, this mimics a "dogmatic" profile, namely confident but inflexible, which correlates with severe performance degradation in safety-critical domains (Health $r=-0.34$), ignoring its domain alignment.

In Figure \ref{fig:correlation_comparison}, we compare Pearson correlations between standard MMLU and the rigorous MMLU-Pro, showing that task difficulty acts as a magnifier for specific personality needs. Extraversion transitions from a mild predictor in MMLU ($r\approx0.13$) to a dominant driver in MMLU-Pro (STEM $r=0.43$, Social Science $r=0.48$). We hypothesize that "Machine Extraversion" serves as a functional proxy for verbosity; high-Extraversion models naturally generate elaborated contexts, triggering implicit Chain-of-Thought (CoT) mechanisms required for complex deduction. In contrast, in the Legal domain, the impact of Neuroticism shifts from negligible in MMLU ($-0.06$) to a strong penalty in MMLU-Pro ($-0.35$)6. This confirms that advanced procedural reasoning requires emotional stability, further supporting the "Tool Regime" hypothesis for specialised domains.

To answer RQ2, we find that competence is bimodal: models succeed either by being \textit{highly expressive Generalist Agents} or \textit{highly suppressed Specialist Tools}. Intermediate profiles suffer from Personality Dissonance. Furthermore, we identify a "Suppression Advantage" for complex tasks, where minimising personality traits is more robust than cultivating them.

\section{The Linguistic Origins of Personality}

To explain the mechanism behind the observed trait shifts, we conducted a linguistic analysis over 200k token sample of continued pretraining data. In particular, we identify specific linguistic signals that have the potential to drive personality induction. These signals include: (1) \textit{imperative ratio} (Imperative), which identifies command-like sentences that issue a direct instruction, request, or advice (e.g., "\textit{Fix} the syntax error", base form verb without nominal subject); (2) \textit{Type-Token Ratio} (TTR), the lexical diversity of the corpus; (3) \textit{Sentence Complexity} (Complexity), the average sentence length of the corpus; (4) \textit{Sentiment variance} (sentiment), sentiment stability via standard deviation; (5) \textit{Detachment Index} (Detachment), impersonality by contrasting the frequency of third-person object references against first- and second-person pronouns (See Appendix~\ref{sec:corpus_analysis} for detailed descriptions.).  

\paragraph{Imperative Syntax creates a "Polite Lecturer" rather than an Agent.}
The Scientific Mathematician exhibits the most distinct linguistic profile, yet its personality scores reveal a nuance in how "instruction" is internalized. The corpus possesses the highest Imperative Ratio ($0.121$), nearly double that of other domains. Paradoxically, this did not result in high Extraversion ($2.99$) relative to the base model. Instead, the determining factor is the Negative Detachment Index ($-0.010$), which signals a high frequency of personal pronouns. Unlike the "detached" legal text, mathematical writing frequently employs the "Academic We" (e.g., "We denote..."). This conditions the model to view reasoning as a cooperative social activity, driving High Agreeableness ($3.14$). However, this social constraints is paired with the lowest Type-Token Ratio ($0.050$) and Complexity ($13.17$), resulting in a model that is helpful but lexically rigid. This "Polite Lecturer" persona—agreeable but unimaginative (Openness $2.95$)—fails to generalize in MMLU STEM tasks ($0.496$) because it prioritizes narrow, formulaic cooperation over independent, open-ended inference.

\paragraph{Transactional Syntax enables the "Suppression Advantage."}
The Technical Communicator provides a crucial counterpoint. Like the Mathematician, it has a Negative Detachment Index ($-0.022$, the lowest in the set), indicating a heavy use of `I' and `You'. However, instead of becoming Agreeable, it develops Low Agreeableness ($2.88$) and Low Extraversion ($2.91$). The differentiator is Sentiment Variance ($0.233$, the second highest). The `I/You' of technical forums is not cooperative but transactional and often corrective (e.g., ``You have a syntax error''), marked by frustration and resolution. This linguistic environment forces the model to strip away social niceties, using personal pronouns purely for object-level problem solving. This validates the "Suppression Advantage": by ignoring the social pressure of "I/You" tokens, the model functions as a neutral instrument, achieving superior reasoning performance on MMLU-Pro.

\begin{table}[t]
\centering
\small
\resizebox{0.7\textwidth}{!}{%
\begin{tabular}{l|ccccc}
\toprule
\textbf{Model Variant} & \textbf{Imperative} & \textbf{TTR} & \textbf{Complexity} & \textbf{Sentiment} & \textbf{Detachment} \\ 
\midrule 
Literary Classicist & 0.024 & 0.073 & 17.08 & \textbf{0.244} & \textbf{0.022} \\
Inventive Technologist & 0.046 & 0.079 & 17.51 & 0.185 & 0.001 \\
Patent Strategist & 0.025 & 0.064 & 21.61 & 0.154 & 0.012 \\
Cultural Scholar & 0.008 & \textbf{0.124} & \textbf{22.09} & 0.184 & 0.020 \\
Technical Communicator & 0.052 & 0.069 & 14.85 & 0.233 & -0.022 \\
Business Advisor & 0.061 & 0.082 & 15.32 & 0.214 & -0.023 \\
Health Advisor & 0.009 & 0.084 & 21.15 & 0.170 & 0.004 \\
Scientific Scholar & 0.038 & 0.059 & 17.12 & 0.168 & -0.012 \\
Scientific Mathematician & \textbf{0.121} & 0.050 & 13.17 & 0.176 & -0.010 \\
Legal Analyst & 0.042 & 0.059 & 19.02 & 0.150 & 0.014 \\
Biomedical Expert & 0.008 & 0.074 & 19.67 & 0.166 & 0.000 \\

\bottomrule
\end{tabular}}
\caption{\textbf{Linguistic Signals of Training Corpora.} We analysed 200k-token samples from each domain corpus over five measures to capture the relationships between linguistic nuance and model performance.}
\label{tab:corpus_dna}
\end{table}

\paragraph{Complexity requires Low Entropy to induce Conscientiousness.}
We identify a critical interaction between Sentence Complexity and Lexical Diversity (TTR). Three variants, the Cultural Scholar, Patent Strategist, and Health Advisor, share extreme Sentence Complexity ($>21$ tokens), forcing them to track long-range dependencies. However, their personality outcomes diverge sharply. The Patent Strategist pairs High Complexity ($21.61$) with Low TTR ($0.064$), channelling this structural burden into High Conscientiousness ($3.30$). In contrast, the Cultural Scholar pairs High Complexity ($22.09$) with High TTR ($0.124$). The combination of structural density and high lexical entropy appears to overwhelm the model's ordering capacity, resulting in Low Conscientiousness ($2.93$). This suggests that ``Machine Conscientiousness", the ability to follow strict rules, emerges only when syntactic complexity is stabilized by a repetitive, predictable vocabulary (Low Entropy).

\section{Conclusion}
This work demonstrates that ``experiences build character" in LLMs: domain exposure induces stable personality profiles that fundamentally dictate problem-solving behaviour. We find that model competence is polarised between Expressive Generalists (e.g., Llama-3-8B) and Suppressed Specialists (e.g., Technical Communicator), with intermediate profiles suffering from performance degradation. For complex analytical tasks, we observe a "Suppression Advantage," where reduced social traits correlate with improved reasoning. Furthermore, our linguistic analysis confirms that these personalities are controllable artefacts of training data, mapping syntax signals to personality traits. Consequently, we propose shifting from reactive prompt engineering to proactive Personality Engineering, curating the linguistic "DNA" of pretraining data to deterministically engineer the cognitive profiles required for specific applications. However, these findings are based on a single base model (Llama-3-8B), and it remains to be verified whether the observed personality dynamics whether the same personality-performance relationships generalise to larger models or those trained on different base corpora. 



\bibliography{colm2026_conference}
\bibliographystyle{colm2026_conference}

\appendix
\section{Appendix}
\begin{table}[t]
\centering
\renewcommand{\arraystretch}{1.05}
\resizebox{\linewidth}{!}{
\begin{tabular}{p{2.7cm}|p{9.5cm}}
\toprule
\textbf{Dataset} & \textbf{Description} \\
\midrule
\textit{ArXiv} & Scientific preprints across STEM disciplines. \\
\textit{PhilPapers} & Philosophical literature and academic writing. \\
\textit{NIH Exporter} & Biomedical research funding abstracts. \\
\textit{DM Mathematics} & Theoretical and applied mathematics papers. \\
\textit{FreeLaw} & U.S. legal case documents and opinions. \\
\textit{Gutenberg} & Classic and public-domain literary works. \\
\textit{PubMed Central} & Full-text biomedical and clinical research. \\
\textit{Wikipedia} & Encyclopaedic general-knowledge text. \\
\textit{Enron Emails} & Corporate and Business Communication. \\
\textit{StackExchange} & Community technical Q\&A discussions. \\
\textit{HackerNews} & Technology and startup discussion threads. \\
\textit{GitHub} & Open-source code, documentation, and developer discourse. \\
\midrule
\multicolumn{2}{l}{\textbf{Roles and corresponding datasets}} \\
\midrule
\end{tabular}
}
\resizebox{\linewidth}{!}{
\begin{tabular}{p{4.3cm}|p{8cm}}
\textit{Scientific Scholar} & ArXiv; PhilPapers; NIH Exporter \\
\textit{Scientific Mathematician} & ArXiv; DM Mathematics \\
\textit{Legal Analyst} & FreeLaw \\
\textit{Literary Classicist} & Gutenberg \\
\textit{Biomedical Expert} & NIH Exporter; PubMed Central \\
\textit{Health Advisor} & PubMed Central; Wikipedia \\
\textit{Business Advisor} & Enron Emails; StackExchange; HackerNews \\
\textit{Technical Communicator} & StackExchange; GitHub \\
\textit{Cultural Scholar} & Gutenberg; Wikipedia \\
\textit{Patent Strategist} & PhilPapers; FreeLaw \\
\textit{Inventive Technologist} & HackerNews; GitHub \\
\bottomrule
\end{tabular}
}
\caption{Overview of domain-specific datasets (top) and their corresponding use by character (role) variants (bottom) in continued pretraining.}
\label{tab:datasets_roles_vertical}
\end{table}

\begin{table*}[t]
\centering
\small
\renewcommand{\arraystretch}{1.25}
\setlength{\tabcolsep}{5pt}
\begin{tabular}{p{3cm} p{6cm} p{4cm}}
\toprule
\textbf{Domain Group} & \textbf{MMLU Categories} & \textbf{MMLU-Pro Categories} \\
\midrule
\textbf{STEM} &
abstract\_algebra, astronomy, college\_chemistry, college\_computer\_science, college\_mathematics, college\_physics, computer\_security, conceptual\_physics, electrical\_engineering, elementary\_mathematics, formal\_logic, high\_school\_computer\_science, high\_school\_mathematics, high\_school\_physics, high\_school\_statistics, machine\_learning &
Mathematics, Physics, Chemistry, Engineering, Computer Science \\
\hline
\textbf{Health \& Biology} &
anatomy, clinical\_knowledge, college\_biology, college\_medicine, human\_aging, medical\_genetics, nutrition, professional\_medicine, virology &
Biology, Health \\
\hline
\textbf{Social Science} &
business\_ethics, econometrics, high\_school\_macroeconomics, high\_school\_microeconomics, management, marketing, professional\_accounting, sociology, public\_relations &
Economics, Business \\
\hline
\textbf{Law \& Policy} &
international\_law, jurisprudence, professional\_law, security\_studies, us\_foreign\_policy, high\_school\_government\_and\_politics &
Law \\
\hline
\textbf{Humanities} &
philosophy, moral\_disputes, moral\_scenarios, prehistory, high\_school\_us\_history, high\_school\_world\_history, high\_school\_european\_history, world\_religions, global\_facts, logical\_fallacies &
Philosophy, History \\
\hline
\textbf{Psychology} &
high\_school\_psychology, professional\_psychology, human\_sexuality &
Psychology \\
\hline
\textbf{Other} &
miscellaneous &
other \\
\bottomrule
\end{tabular}
\caption{Unified domain grouping between MMLU and MMLU-Pro. MMLU’s 57 fine-grained categories are aggregated into seven high-level domains aligned with MMLU-Pro’s broader taxonomy, enabling consistent domain-level comparison across benchmarks. A dashed line separates STEM, health, and social domains from humanities and behavioural domains.}
\label{tab:mmlu_mapping}
\end{table*}

\section{Linguistic Corpus Analysis}\label{sec:corpus_analysis}
To investigate the causal origins of the observed personality traits, we performed a "Corpus Forensics" analysis on the pretraining data. We quantified the linguistic "DNA" of each domain corpus using \textit{five stylistic metrics}, each serving as a proxy for a specific dimension of the Big Five personality traits:

\begin{enumerate}
    \item \textbf{Imperative Ratio} (Proxy for Extraversion/Assertiveness): This metric measures the frequency of command-like sentence structures. We identify a sentence as imperative if its root token is a verb in the base form (POS tag VERB, morphological tag VB) and it lacks a nominal subject (nsubj) dependency.
    
    \small
    $$\text{Imperative Ratio} = \frac{\text{Count(Imperative Sentences)}}{\text{Total Sentences}}$$
    \normalsize
    \textbf{\textit{Hypothesis}}: High imperative frequency (common in technical documentation and mathematical proofs) conditions the model to generate assertive, direct text, which is interpreted by personality classifiers as high Extraversion.
    \item \textbf{Type-Token Ratio (TTR)} (Proxy for Openness):TTR quantifies the lexical diversity of the corpus. It is calculated as the number of unique word types divided by the total number of word tokens in a sample.

    \small
    $$\text{TTR} = \frac{|V|}{\sum_{w \in V} \text{count}(w)}$$
    \normalsize
    \textit{\textbf{Hypothesis}}: Corpora with high TTR (e.g., literature) expose the model to a broader semantic space, enhancing its "imagination" and resulting in higher Openness scores. Repetitive corpora (e.g., mathematics) lead to lower Openness.
    \item \textbf{Sentence Complexity} (Proxy for Conscientiousness): We define complexity as the average sentence length in tokens. This captures the structural depth and formality of the text.

    \small
    $$\text{Avg. Sentence Length} = \frac{\text{Total Tokens}}{\text{Total Sentences}}$$
    \normalsize
    \textbf{\textit{Hypothesis}}: Long, syntactically complex sentences (common in legal and patent texts) require the model to track long-range dependencies and adhere to strict formal constraints, manifesting downstream as high Conscientiousness and stability.
    \item \textbf{Sentiment Variance} (Proxy for Neuroticism): To estimate emotional stability, we calculated the polarity of every sentence using TextBlob and computed the standard deviation of these scores across the corpus.

    \small
    $$\text{Sentiment Variance} = \sigma(\text{Sentence Polarities})$$
    \normalsize
    \textbf{\textit{Hypothesis}}: A high variance indicates frequent shifts between positive and negative extremes (emotional volatility), serving as a proxy for Neuroticism. Low variance implies a neutral, clinical tone.
    \item \textbf{Detachment Index} (Proxy for Low Agreeableness/Extraversion): This metric quantifies the "impersonality" of the text by contrasting third-person/object references against first/second-person references. We define personal pronouns as $P_{personal} = \{i, me, my, we, us, our, you\}$ and impersonal pronouns as $P_{impersonal} = \{it, its, he, she, they, ...\}$.
    
    \small
    $$\text{Detachment} = \frac{\text{count}(P_{impersonal}) - \text{count}(P_{personal})}{\text{Total Tokens}}$$
    \normalsize
    \textbf{\textit{Hypothesis}}: A high detachment score indicates an objective, third-person perspective (typical of scientific or legal texts), correlating with lower Agreeableness and "social" Extraversion.
\end{enumerate}

\end{document}